\documentclass{article}

\usepackage[utf8]{inputenc}
\usepackage{custom_alias}
\usepackage{amsmath, amssymb, amsthm}
\usepackage{breakcites}
\usepackage{color}
\usepackage{subcaption}
\usepackage{graphicx}
\usepackage{hyperref}
\usepackage{authblk}
\usepackage{booktabs}
\usepackage[parfill]{parskip}
\usepackage{multirow}
\usepackage[accepted]{icml2019}
\DeclareMathOperator*{\argmin}{argmin}

\usepackage{eqparbox}

\begin{document}

\twocolumn[
\icmltitle{Improving Evolutionary Strategies with Generative Neural Networks}

\begin{icmlauthorlist}
\icmlauthor{Louis Faury}{equal,to}
\icmlauthor{Cl\'ement Calauz\`enes}{equal}
\icmlauthor{Olivier Fercoq}{to}
\icmlauthor{Syrine Krichen}{equal}
\end{icmlauthorlist}

\icmlaffiliation{equal}{Criteo AI Labs, 32 Rue Blanche, Paris, France}
\icmlaffiliation{to}{LTCI, Télécom ParisTech, Université Paris Saclay, France}

\icmlcorrespondingauthor{Louis Faury}{l.faury@criteo.com}

\icmlkeywords{Machine Learning, ICML, Evolutionary Strategies, Generative Neural Networks}

\vskip 0.3in
]

\printAffiliationsAndNotice{}

\begin{abstract}
    Evolutionary Strategies (ES) are a popular family of black-box zeroth-order optimization algorithms which rely on search distributions to efficiently optimize a large variety of objective functions. This paper investigates the potential benefits of using highly flexible search distributions in classical ES algorithms, in contrast to standard ones (typically Gaussians). We model such distributions with Generative Neural Networks (GNNs) and introduce a new training algorithm that leverages their expressiveness to accelerate the ES procedure. We show that this tailored algorithm can readily incorporate existing ES algorithms, and outperforms the state-of-the-art on diverse objective functions. 
\end{abstract}

\section{Introduction}

We are interested in the global minimization of a black-box objective function, only accessible through a \emph{zeroth-order} oracle. In many instances of this problem the objective is expensive to evaluate, rendering brute force optimization undesirable. Moreover it is possibly non-convex and potentially highly multi-modal, hence its global optimization cannot be done greedily but requires a careful balance between exploitation and exploration of the \emph{optimization landscape} (the surface defined by the objective).

The family of algorithms used to tackle such a problem is usually dictated by the \emph{cost} of one evaluation of the objective function - or equivalently by the maximum number of function evaluations that are reasonable to make, and by the precision requirement. For instance, Bayesian Optimization \cite{jones1998efficient, shahriari2016taking} targets problems of very high evaluation cost, where the global minimum must be approximately discovered after a few hundreds of function evaluations. When aiming for a better precision and hence having a  larger budget (typically several thousands of function evaluations), a popular class of algorithms is the one of Evolutionary Strategies (ES) \cite{rechenberg1978evolutionsstrategien, schwefel1977numerische}, a family of heuristic search procedures.

ES algorithms rely on a \emph{search distribution}, which role is to propose queries of potentially small value of the objective function. This search distribution is almost always  chosen to be a multivariate Gaussian. It is namely the case of the Covariance Matrix Adaptation Evolution Strategies (CMA-ES) \cite{hansen2001completely}, a state-of-the-art ES algorithm made popular in the machine learning community by its good results on hyper-parameter tuning \cite{friedrichs2005evolutionary, loshchilov2016cma}. It is also the case for Natural Evolution Strategies (NES) \cite{wierstra2008natural} algorithms, which were recently used for direct policy search in Reinforcement Learning (RL) and shown to compete with state-of-the-art MDP-based RL techniques \cite{salimans2017evolution}. Occasionally, other distributions have been used; e.g. fat-tails distributions like the Cauchy were  shown to over-perform the Gaussian for highly multi-modal objectives \cite{schaul2011high}.

We argue in this paper that in ES algorithms, the choice of a given parametric distribution (Gaussian, Cauchy, ..) constitutes a \emph{potentially harmful implicit constraint} for the stochastic search of a global minimum.  To overcome the limitations of classical parametric search distributions, we propose using \emph{flexible} distributions generated by bijective Generative Neural Networks (GNNs), with computable and differentiable log-probabilities. We discuss why common existing optimization methods used by ES algorithms cannot be directly used to train such models and design a tailored algorithm that efficiently train GNNs for an ES objective. We show how this new algorithm can readily incorporate existing ES algorithms that operates on simple search distributions, like the Gaussian. Finally, we show that this algorithm outperforms state-of-the-art ES algorithms on a variety of objective functions.

%\paragraph{Related work} Evolutionary Strategies \cite{rechenberg1978evolutionsstrategien, %schwefel1977numerische} are age-old algorithms, which are still undergoing innovations and improvements %\cite{conti2018improving}. 

%Using GNNs for randomized optimization was already investigated for differentiable objective %\cite{faury2018neural, lopez2018easing}, but to the best our our knowledge it is the first time they are %used for zeroth-order optimization. 

We introduce the problem and provide background on Evolutionary Strategies in Section \ref{sec::prel}. We discuss the role of GNN in generating flexible search distributions in Section \ref{sec::gnn}. We explain why usual algorithms fail to train GNNs for an ES objective and introduce a new algorithm in Section \ref{sec::algo}. Finally we report experimental results in Section \ref{sec::exp}. 

\section{Preliminaries}
\label{sec::prel}

In what follows, the objective function will be noted $f$ and we will consider its global optimization over a compact set $\mathcal{X}\subset\mathbb{R}^d$:
\begin{align}
    x^* \in \argmin_{x\in\mathcal{X}} f(x)
\end{align}
The symbol $\pi$ will generically denote a probability density function over $\mathcal{X}$.

\subsection{Evolutionary Strategies}
The generic procedure followed by ES algorithms is presented in Algorithm~\ref{alg::es}. To make the update step tractable, the search distribution is tied to a family of distributions and parametrized by a real-valued parameter vector $\theta$ (e.g. the mean and covariance matrix of a Gaussian), and would therefore now be referred to as $\pi_\theta$. This update step constitute the main difference between ES algorithms.

\begin{algorithm}[tb]
   \caption{Generic ES procedure}
   \label{alg::es}
\begin{algorithmic}
   \STATE {\bfseries Input:} objective $f$, distribution $\pi_0$, population size $n$
   \REPEAT
   \STATE \emph{(Sampling)} Sample
   $ x_1, \hdots, x_n \overset{\text{i.i.d}}{\sim} \pi_t$
  \STATE \emph{(Evaluation)} Evaluate $f(x_1), \hdots, f(x_n)$.
  \STATE  \emph{(Update)} Update $\pi_t$ to produce $x$ of potentially smaller objective values. 
   \UNTIL{convergence}
\end{algorithmic}
\end{algorithm}

\paragraph{Natural Evolution Strategies} One principled way to perform that update is to minimize the expected objective value over samples $x$ drawn from $\pi_\theta$:
\begin{align}
    J(\theta) \triangleq \lE_{\pi_\theta} \left[f(x)\right]
    \label{eq::es_objective}
\end{align}
When the search distribution is parametric and tied to a parameter $\theta$, this objective can be differentiated with respect to $\theta$ thanks to the log-trick:
\begin{align}
 \frac{\partial J(\theta)}{\partial \theta} = \lE_{\pi_{\theta}} \left[f(x)\frac{\partial \log \pi_\theta(x)}{\partial \theta} \right]
    \label{eq::es_objective_grad}
\end{align}
This quantity can be approximated from samples  - it is known as the score-function or REINFORCE \cite{williams1992simple} estimator, and provides a direction of update for $\theta$. Unfortunately, naively following a stochastic version of the gradient \eqref{eq::es_objective_grad} -- a procedure called Plain Gradient Evolutionary Strategies (PGES) -- is known to be highly ineffective. PGES main limitation resides in its instability when the search distribution is concentrating, making it unable to \emph{precisely} locate any local minimum. To improve over the PGES algorithm the authors of \cite{wierstra2008natural} proposed to descend $J(\theta)$ along its \emph{natural gradient} \cite{amari1998natural}. More precisely, they introduce a trust-region optimization scheme to limit the instability of PGES, and minimize a linear approximation of $J(\theta)$ under a Kullback-Leibler (KL) divergence constraint: 
\begin{equation}
\begin{aligned}
   \argmin_{\delta\theta}& \quad J(\theta+\delta\theta) \simeq J(\theta) +\delta\theta^T\nabla_\theta J(\theta) \\
    &\text{s.t} \quad \text{KL}(\pi_{\theta+\delta\theta} \vert \vert \pi_\theta)\leq \epsilon
\end{aligned}
\label{eq::nes_tr}
\end{equation}
To avoid solving analytically the trust region problem \eqref{eq::nes_tr}, \cite{wierstra2008natural} shows that its solution can be approximated by:
\begin{align}
    \delta\theta^* \propto -F_\theta^{-1} \nabla_\theta J(\theta)
    \label{eq::nes_descent}
\end{align}
where 
\begin{align} 
F_\theta = \lE_{\pi_\theta}\left[\nabla_\theta \log\pi_\theta(x)\nabla_\theta \log\pi_\theta(x)^T\right]
\end{align}is the Fischer Information Matrix (FIM) of $\pi_\theta$. The parameter $\theta$ is therefore not updated along the negative gradient of $J$ but rather along $F_\theta^{-1} \nabla_\theta J(\theta)$, a quantity known as the natural gradient. The FIM $F_\theta$ is known analytically when $\pi_\theta$ is a multivariate Gaussian and the resulting algorithm, Efficient Natural Evolutionary Strategies (xNES) \cite{sun2009efficient} has been shown to reach state-of-the-art performances on a large ES benchmark.

\paragraph{CMA-ES} Naturally, there exist other strategies to update the search distribution $\pi_\theta$. For instance, CMA-ES relies on a variety of heuristic mechanisms like covariance matrix adaptation and evolution paths, but is only defined when $\pi_\theta$ is a multivariate Gaussian. Explaining such mechanisms would be out of the scope of this paper, but the interested reader is referred to the work of \cite{hansen2016cma} for a detailed tutorial on CMA-ES.

\subsection{Limitations of classical search distributions} 
\label{subsec::limit}
ES implicitly balances the need for exploration and exploitation of the optimization landscape. The exploitation phase consists in updating the search distribution, and exploration happens when samples are drawn from the search distribution's tails. The key role of the search distribution is therefore to produce a support adapted to the landscape's structure, so that new points are likely to improve over previous samples.

\begin{figure}
    \centering
    \includegraphics[width=0.9\linewidth]{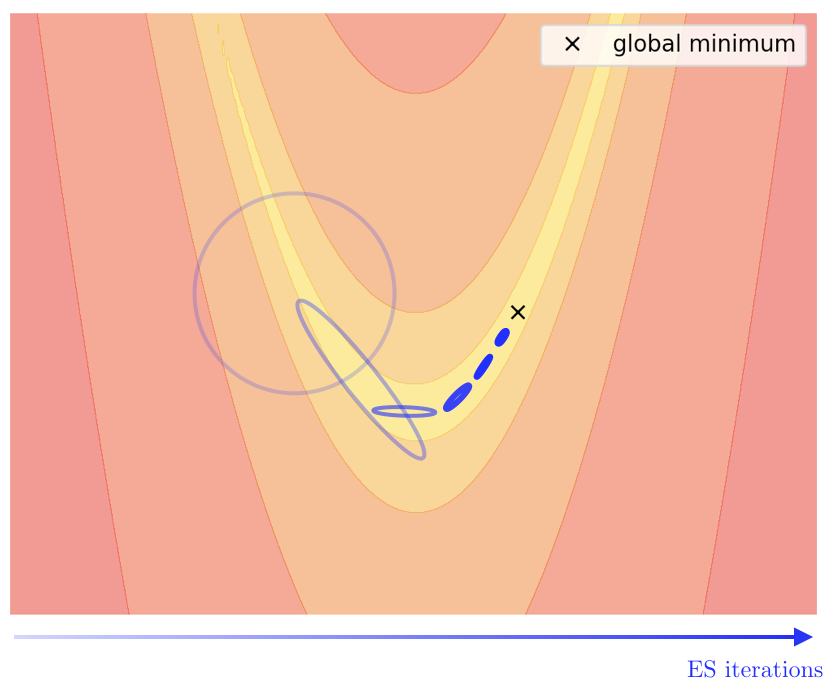}
    \caption{Example of an undesirable behavior of a Gaussian search distribution trained by xNES on the Rosenbrock function. The solid lines represent the one standard-deviation isoline of the distribution, from the beginning of the algorithm (shaded) until it reaches the straight part of the valley (full color).}
    \label{fig::rosenbrock_example}
\end{figure}

We argue here that the choice of a given parametric distribution (the multivariate Gaussian distribution being overwhelmingly represented in state-of-the-art ES algorithms) constitutes a \emph{potentially harmful implicit constraint} for the stochastic search of a global minimum. For instance, any symmetric distribution will be slowed down in curved valleys because of its inability to continuously curve its density. This lack of flexibility will lead it to drastically reduce its entropy, until the curved valley looks \emph{locally} straight. At this point, the ES algorithm resembles a hill-climber and barely takes advantage of the exploration abilities of the search distribution. An illustration of this phenomenon is presented in Figure \ref{fig::rosenbrock_example} on the Rosenbrock function, a curved valley with high-conditioning. Another limitation of classical search distribution is their inability to follow \emph{multiple hypothesis}, that is to explore at the same time different local minima. Even if mixture models can achieve that flexibility, hyper-parameters like the number of mixtures have optimal values that are impossible to guess \emph{a priori}. 

We want to introduce \emph{flexible} search distributions to overcome these limitations. Such distributions should, despite their expressiveness, be easily trainable. We should also be concerned when designing them with their role in the exploration / exploitation trade off: a search distribution with too much capacity could over-fit some seemingly good samples, leading to premature convergence. To sum-up, we want to design search-distributions that are:
\begin{itemize}
    \item more flexible than classical distributions
    \vspace{-0.5em}\item yet easily trainable
    \vspace{-0.5em}\item while keeping control over  exploration / exploitation
\end{itemize}

In the following section, we carefully investigate the class of Generative Neural Networks (GNNs) to find a parametric class of distributions satisfying such properties. 

\section{Flexible search distributions with GNNs}
\label{sec::gnn}

Generative Neural Networks \cite{mackay1995bayesian} have been studied in the context of \emph{density estimation} and shown to be able to model complex and highly multimodal distributions \cite{srivastava2017veegan}. 
We propose here to leverage their expressiveness for ES, and train them in a principled way thanks to the ES objective:
$$
J(\pi) = \lE_{\pi}\left[f(x)\right]
$$
As discussed in Section \ref{sec::prel}, optimizing $J(\pi)$ with gradient-based methods is possible through the score-function estimator, which requires to be able to compute and efficiently differentiate the log-probabilities of $\pi$.

\subsection{GNN backgound}
The core idea behind a GNN is to map a \emph{latent} variable $z\in\mathcal{Z}$ drawn from a known distribution $\nu_\mu$ to an output variable $ x = g_\eta(z)$ where $g_\eta$ is the forward-pass of a neural network. The parameter $\eta$ represents the weights of this neural network while $\mu$ describe the degrees of freedom of the latent space distribution $\nu_\mu$. We denote $\theta \!=\! (\mu, \eta)$ and $\pi_\theta(x)$ the density of the resulting output variable $x$.

For general neural network architectures, it is impossible to compute the likelihood $\pi_\theta(x)$ of samples $x$ drawn from the GNN. This is namely why their are often trained with adversarial methods \cite{goodfellow2014generative} for sample generation purposes, bypassing the need of computing densities, but at the expense of a good density estimation (mode-dropping). An alternative to adversarial methods was proposed with Variational Auto-Encoders \cite{kingma2013auto} however at the cost of learning two neural networks (an encoder and a decoder). A less computationally expensive method consists in restricting the possible architectures to build \emph{bijective} GNNs, which allows the exact computation of the distribution's density.  

Indeed,  if $g_\eta$ is a bijection from $\cZ$ to $\cX$ with inverse $h_\eta \triangleq g_\eta^{-1}$, the change of variable formula provides a way to compute $\pi_\theta(x)$:
\begin{align}
    \pi_\theta(x) = \nu_\mu(h_\eta(x)) \cdot \left\vert \frac{\partial h_\eta(x)}{\partial x}\right\vert
\label{eq::inverse_transform}
\end{align}
To have a tractable density one therefore needs to ensure that the determinant of the Jacobian $\left\vert \partial h_\eta(x)/\partial x\right\vert$ is easily computable. Several models satisfying these two properties (\emph{i.e} bijectivity and computable Jacobian) have been proposed for density estimation \cite{rippel2013high, dinh2014nice, dinh2016density}, and proved their high expressiveness despite their relatively simple structure. 

Bijective GNNs therefore answer two of our needs when building our new search distribution: flexibility and easiness to train. In this work, we will focus on one bijective GNN, the Non-Linear Independent Component Estimation \cite{dinh2014nice} (NICE) model, for its numerical stability and \emph{volume preserving} properties.

\subsection{NICE model}
The authors of NICE proposed to build complex yet invertible transformations through the use of \emph{additive coupling layers}. An additive coupling layer leaves half of its input unchanged, and adds a non-linear transformation of the first half to the second half. More formally, by noting $v=[v_1, v_2]$ the output of a coupling layer and $u=[u_1, u_2]$ its input, one has:
\begin{equation}
\begin{aligned}
    v_1 &= u_{1}\\
    v_2 &= u_2 + t(u_1)
\end{aligned}
\end{equation}
where $t$ is an arbitrarily complex transformation - modelled by a Multi-Layer Perceptron (MLP) with learnable weights and biases. This transformation has unit Jacobian and is easily invertible:
\begin{equation}
\begin{aligned}
u_1 &= v_1\\
    u_2 &= v_2 - t(v_1)
\end{aligned}
\end{equation}
and only requires a feed-forward pass on the MLP $t$. 
The choice of the decomposition $u=[u_1, u_2]$ can be arbitrary, and is performed by applying a binary filter to the input. By stacking additive coupling layers, one can create complex and possibly multimodal distributions. The inversion of the resulting network is independent of the complexity of $t$. Also, as noted by the authors of \cite{dinh2014nice}, it is enough to stack only three coupling layers with complementary binary masks in order to make sure that every dimension influences all the others. The density of the resulting distribution is then readily computable thanks to the inverse transform theorem \eqref{eq::inverse_transform}. 

\subsection{Volume preserving properties}
The transformation induced by NICE is \emph{volume preserving} (it has a unitary Jacobian determinant). This is quite desirable in a ES context, as the role of concentrating the distribution on a minimum can be left to the latent space distribution $\nu_\mu$. The role of the additive coupling layers is therefore only to introduce non-linearities in the inverse transform $h_\eta$ so that the distribution is better adapted to the optimization landscape. The fact that this fit can be done while preserving the global volume of the distribution forces the model to align its tails with the optimization landscape, which is likely to improve the quality of future exploration steps. The NICE model therefore fits perfectly our needs for a flexible search distribution that is easy to train, and that provides enough control on the exploration / exploitation trade-off. Indeed, other bijective GNN models like the Real-NVP \cite{dinh2016density} introduce non-volume preserving transformations, which can easily overfit on good samples and lead to premature convergence. 

\section{An efficient training algorithm}
\label{sec::algo}
We are now equipped with enough tools to use GNNs for ES: an adapted model (NICE) for our search distribution $\pi_\theta$, and an objective to train it with:
\begin{align}
    J(\theta) = \mathbb{E}_{\pi_\theta}\left[f(x)\right]
    \label{eq::es_objective_2}
\end{align}
Here, $\theta$ describes \emph{jointly} the free parameters of the latent distribution $\nu_\mu$ and $\eta$, the weights and biases of the MLPs forming the additive coupling layers.

We start this section by explaining why existing training strategies based on the objective \eqref{eq::es_objective_2} are not sufficient to truly leverage the flexibility of GNNs for ES, before introducing a new algorithm tailored for this task. 

\subsection{Limitations of existing training strategies}

We found that the PGES algorithm (naive stochastic gradient descent of \eqref{eq::es_objective_2} with the score-function estimator \eqref{eq::es_objective_grad}) applied to the NICE distribution suffers from the same limitations as when applied to the Gaussian - that is inability to precisely locate any local minimum. As for the Gaussian, training the NICE distribution for ES requires employing more sophisticated algorithms - such as NES. 

Using the natural gradient for the GNNs distributions is not trivial. First the Fischer Information Matrix $F_\theta$ is not known analytically and must be estimated via Monte-Carlo sampling, thereby introducing approximation errors.  Also, we found that the approximations justifying to follow the descent direction provided by the natural gradient are not adapted to the NICE distribution. Indeed, the assumption behind the NES update \eqref{eq::nes_descent} is that the loss $J(\theta)$ can be (locally) well approximated by the quadratic objective:
\begin{align}
    J(\theta+\delta\theta) = J(\theta) +\delta\theta^T\nabla_\theta J(\theta) + \frac{\lambda}{2}\delta\theta ^T F_\theta \delta \theta
    \label{eq::quad_obj}
\end{align}
where $\lambda$ is a given non-negative Lagrange multiplier. For NICE, given the highly non-linear nature of $\pi_\theta$ this approximation is bound to fail even close to the current parameter $\theta$ and will lead to spurious updates. A classical technique \cite{NoceWrig06, martens2010deep} to avoid such updates is to artificially increase the curvature of the quadratic term, and is known as \emph{damping}. Practically, this implies using $F_\theta+\beta I$ instead of $F_\theta$ as the local curvature metric, with $\beta$ a non-negative damping parameter. 

We found in our experiments that to ensure continuous decrease of $J(\theta)$, and because of its highly non-linear nature when using the GNNs, the damping parameter $\beta$ has to be set to such high values that the modifications in the search distributions are too small to quickly make progress and by no means reaches state-of-the-art performances. We observed that even if the training of the additive coupling layers is performed correctly (i.e the distribution has the correct \emph{shape}), it is high damping of the latent space parameters that prevents the distribution to quickly concentrate when a minimum is found. 

It is unclear how the damping parameter should be adapted to avoid spurious update, while still allowing the distribution to make large step in the latent space and ensure fast concentration when needed. Hence, in the following, we rather present an \emph{alternated minimization} scheme to bypass the issues raised by natural gradient training for GNN distributions in a ES context. 

\subsection{Alternated minimization}
\paragraph{Latent space parameters optimization} So far, we used the parameter $\theta$ to describe both the free parameters $\mu$ of the latent space distribution $\nu_\mu$ and the ones of the non-linear transformations of the additive coupling layers $\eta$, and the optimization of all these parameters was done jointly. Separating the roles of $\mu$ and $\eta$, the initial objective \eqref{eq::es_objective} can be rewritten as follows:
\begin{align}
        J(\theta) &= J(\mu, \eta)\\
                  &= \lE_{z\sim\nu_\mu}\left[ f(g_{\eta}(z))\right]
    \label{eq::latent_repres}
\end{align}
This rewriting of the initial objective can give us a new view of the role of GNNs in ES algorithms. Namely, that this role can be to learn an efficient transformation $g_\eta$ so that the representation of $f\circ g_\eta$ in the latent space is \emph{easy} to minimize for the latent distribution. If $\nu_\mu$ is a standard distribution (i.e efficiently trainable with the natural gradient) and that $f\circ g_{\eta}$ is a \emph{well structured} function (i.e one for which $\nu_\mu$ is an efficient search distribution), then the single optimization of $\mu$ by classical methods should avoid the limitations discussed earlier. This new representation motivates us to design a new training algorithm that now optimizes the parameters $\mu$ and $\eta$ \emph{separately}. 

\paragraph{Alternated Minimization} In the following, we will replace the notation $\pi_\theta$ with $\pi_{\mu, \eta}$ to refer to the NICE distribution with parameter $\theta=(\mu, \eta)$. We want to optimize $\mu$ and $\eta$ in an alternate fashion, which means performing the following updates at every step of the ES procedure presented in Algorithm~\ref{alg::es}:
\begin{subequations}
\begin{align}
    \mu_{t+1} &= \argmin_\mu J(\mu, \eta_t) \label{mu_update}\\
     \eta_{t+1} &= \argmin_\eta J(\mu_{t+1}, \eta) \label{eta_update}
\end{align}
\end{subequations}

This means that at iteration $t$, samples will be drawn from $\pi_{\mu_t, \eta_{t}}$ and will serve to first optimize the latent space distribution parameters $\mu$, and then the additive coupling layers parameters $\eta$. The next population will therefore be sampled with $\pi_{\mu_{t+1}, \eta_{t+1}}$.

The update \eqref{mu_update} of the latent space parameters is naturally derived from the new representation \eqref{eq::latent_repres} of the initial objective. Indeed, $\mu$ can be updated via natural gradient ascent of $J(\mu, \eta_t)$ - that is with keeping $\eta=\eta_t$ fixed. Practically, this therefore reduces to applying a NES algorithm to the latent distribution $\nu_{\mu}$ on the modified objective function $f\circ g_{\eta_t}$. 

Once the latent space parameters updated, the coupling layers parameters should be optimized with respect to:
\begin{align}
    J(\mu_{t+1}, \eta) = \lE_{\pi_{\mu_{t+1}, \eta}}\left[ f(x) \right]
    \label{eq::eta_optim}
\end{align}

Because the available samples are drawn from $\pi_{\mu_t, \eta_t}$, the objective \eqref{eq::eta_optim} must be unbiased with \emph{importance sampling} if we wish to approximate it from samples:
\begin{align}
     J(\mu_{t+1}, \eta) = \lE_{\pi_{\mu_t, \eta_t}} \left[ f(x) \frac{\pi_{\mu_{t+1}, \eta}(x)}{\pi_{\mu_t, \eta_t} (x)}\right]
     \label{eq::eta_optim_off}
\end{align}
The straightforward minimization of the \emph{off-line} objective \eqref{eq::eta_optim_off} is known to lead to degeneracies \cite{swaminathan2015counterfactual}, and must therefore be regularized. For our application, it is also desirable to make sure that the update $\eta$ does not undo the progress made in the latent space - or in other words, we want to regularize the change in $f\circ g_{\eta}$. To that extent, we add a trust-region to the minimization of \eqref{eq::eta_optim_off} to ensure that the KL divergence between $\pi_{\theta_{t+1}}$ and $\pi_{\mu_{t+1}, \eta_t}$ remains bounded by a small quantity $\varepsilon$:
\begin{equation}
    \begin{aligned}
        \eta_{t+1} = \argmin_\eta \quad & \lE_{\pi_{\theta_t}} \left[ f(x) \frac{\pi_{\mu_{t+1}, \eta}(x)}{\pi_{\mu_t, \eta_t} (x)}\right]\\
        & \text{s.t} \quad \text{KL}\left(\pi_{\mu_{t+1}, \eta_t} \vert \vert  \pi_{\mu_{t+1}, \eta} \right)\leq \varepsilon
    \end{aligned}
    \label{eq::eta_optim_tr}
\end{equation}
This problem can be efficiently approximated by a simple penalization scheme:
\begin{equation}
\begin{aligned}
   \lE_{\pi_{\theta_t}} &\left[ f(x) \frac{\pi_{\mu_{t+1}, \eta}(x)}{\pi_{\mu_t, \eta_t} (x)}\right] \\ &+\lambda \text{KL}\left(\pi_{\mu_{t+1}, \eta_t} \vert \vert  \pi_{\mu_{t+1}, \eta} \right)
    \end{aligned}
    \label{eq::eta_optim_regul}
\end{equation}

where $\lambda$ is a non-negative penalization coefficient. Its value can be adapted through a simple scheme presented in \cite{schulman2017proximal} to ensure that the constraint arising in \eqref{eq::eta_optim_tr} is satisfied. The KL divergence can readily be approximated by Monte-Carlo sampling of $\pi_{\mu_{t+1}, \eta_t}$ (which only requires sampling in the distribution) and a minimum of \eqref{eq::eta_optim_regul} found by a gradient-based algorithm.  We found the initial value of $\lambda$ has little practical impact, as it is quickly adapted through the optimization procedure. 

To sum up, we propose optimizing the latent distribution and the coupling layers separately. The latent space is optimized by natural gradient descent, and the coupling layers via an off-policy objective with a Kullback-Leibler divergence penalty. We call this algorithm GNN-ES for Generative Neural Networks Evolutionary Strategies. 

\paragraph{Latent space optimization}
It turns out the GNN-ES can be readily modified to incorporate virtually \emph{any} existing ES algorithms that operates on the simple distribution $\nu_\mu$. For instance, if $\nu_\mu$ is set to be a multivariate Gaussian with learnable mean and covariance matrix, the latent space optimization \eqref{mu_update} can be performed by either xNES or CMA-ES. This holds for any standard distribution $\nu_\mu$ and any ES algorithm operating on that distribution. 

This remark allows us to place GNN-ES in a more general framework and to understand it as a way to improve currently existing ES algorithm, by providing a principled way to learn complex, non-linear transformations on top of rather standard search distributions (like the Gaussian). In what follows, we will use the GNN prefix in front of existing ES algorithm to describe their augmented version with our algorithm. 

This new way of understanding GNN-ES highlights the need for a volume-preserving transformation, like the one provided by NICE. Indeed, the role of non-linearities is now only to provide a better representation of the objective function in the latent space \emph{without} changing the volume, leaving the exploration / exploitation balance to the ES algorithm dedicated to training the latent distribution. 

\paragraph{Algorithm overview}
We provide in Algorithm \ref{alg::gnnes} the pseudo-code for the generic algorithm GNN-$\mathcal{A}$-ES. Its principal inputs are the nature of the latent distribution $\nu_\mu$  and the ES algorithm $\mathcal{A}$ used to optimize it. 

\begin{algorithm}[tb]
   \caption{GNN-$\mathcal{A}$-ES (ex: GNN-xNES, GNN-CMA-ES)}
   \label{alg::gnnes}
\begin{algorithmic}
   \STATE {\bfseries Input:} objective $f$, distribution $\nu$ and its related ES algorithm $\mathcal{A}$, initial parameter $\theta_0 = (\mu_0, \eta_0)$, initial $\lambda_0$, radius $\varepsilon$, population size $N$, KL sample size $M$. 
   \REPEAT
   \STATE \emph{Sampling} 
   \STATE Sample $Z = z_1, \hdots z_N \overset{\text{i.i.d}}{\sim} \nu_{\mu_t}$.
   \STATE Feed-forward on $g_{\eta_t}$ to obtain $ x_1, \hdots x_N \overset{\text{i.i.d}}{\sim} \pi_{\theta_t}$.
   \STATE \emph{Evaluation} Evaluate $F=f(x_1), \hdots, f(x_N)$.
   \STATE \emph{Latent space optimization}
    $$\mu_{t+1} \leftarrow \mathcal{A}\left(\mu_t,(Z,F)\right)$$
    \COMMENT{apply $\mathcal{A}$ to the latent distribution}
    \STATE \emph{Bijective network optimization}
    \STATE Sample $\tilde{x}_1, \hdots, \tilde{x}_M \overset{\text{i.i.d}}{\sim} \pi_{\mu_{t+1},\eta_t}$.
    \STATE Compute $$\tilde{KL}(\mu_t, \eta) \leftarrow \frac{1}{M}\sum_{i=1}^M \log\left(\frac{\pi_{\mu_{t+1}, \eta_t}(\tilde{x}_i)}{\pi_{\mu_{t+1}, \eta}(\tilde{x}_i)}\right)$$
    \STATE Minimize w.r.t $\eta$:
    $$
        \frac{1}{N}\sum_{i=1}^N f(x_i) \frac{\pi_{\mu_{t+1}, \eta}(x_i)}{\pi_{\mu_{t+1}, \eta_t}(x_i)} + \lambda_t \tilde{KL}(\mu_t, \eta) 
    $$
    to obtain $\eta_{t+1}$.
    \STATE \emph{Adapting $\lambda$}
    \IF{$\tilde{KL}(\mu_t, \eta_{t+1}) > 2\varepsilon$}
    \STATE $\lambda_{t+1} \leftarrow 1.5\lambda_t$
    \ELSIF{$\tilde{KL}(\mu_t, \eta_{t+1}) < 0.5\varepsilon$}
    \STATE $\lambda_{t+1} \leftarrow \lambda_t/1.5$
    \ENDIF
    
   \UNTIL{convergence}
\end{algorithmic}
\end{algorithm}

\paragraph{Using historic data}
Because the objective \eqref{eq::eta_optim_off} uses importance sampling and therefore is off-line by nature, any historic data can be used to estimate its left-hand side. Therefore, all previously sampled points can be stored in a buffer and used to optimize the coupling layers parameters. On the optimization landscapes we considered, we didn't find this technique to bring a significant uplift to GNN-ES performances. In-depth exploration of this augmentation, popular in similar settings like MDP-based Reinforcement Learning and counterfactual reasoning \cite{nedelec2017comparative, agarwal2017effective}, was therefore left to future work.

\section{Experimental results}
\label{sec::exp}
\subsection{Visualization}
We present here two-dimensional visualizations of the behavior of a GNN distribution trained with GNN-xNES versus a Gaussian distribution trained by xNES. 

\begin{figure}
\begin{subfigure}{0.49\linewidth}
    \centering
    \includegraphics[width=\linewidth]{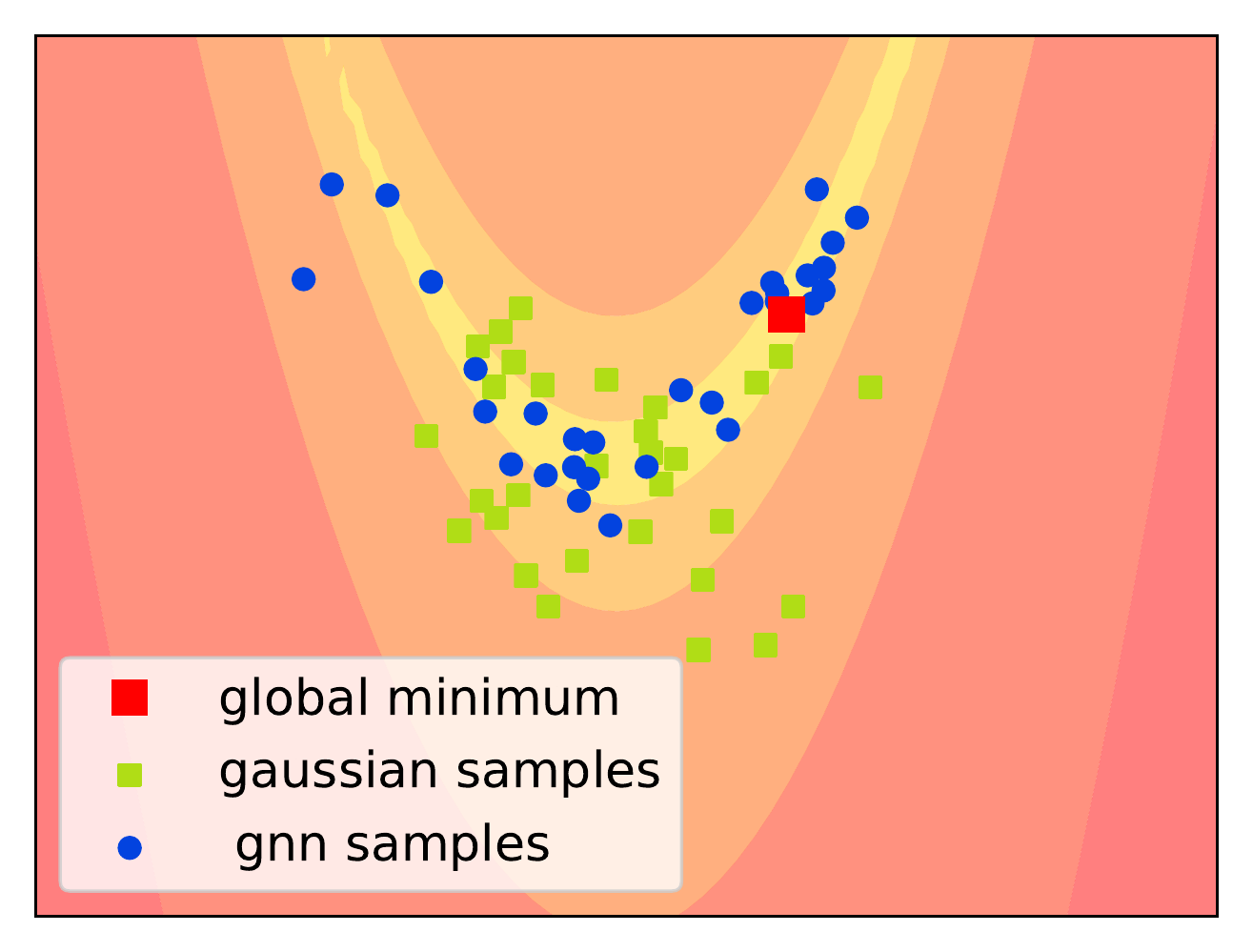}
    \caption{Iteration 5}
\end{subfigure}
\begin{subfigure}{0.49\linewidth}
    \centering
    \includegraphics[width=\linewidth]{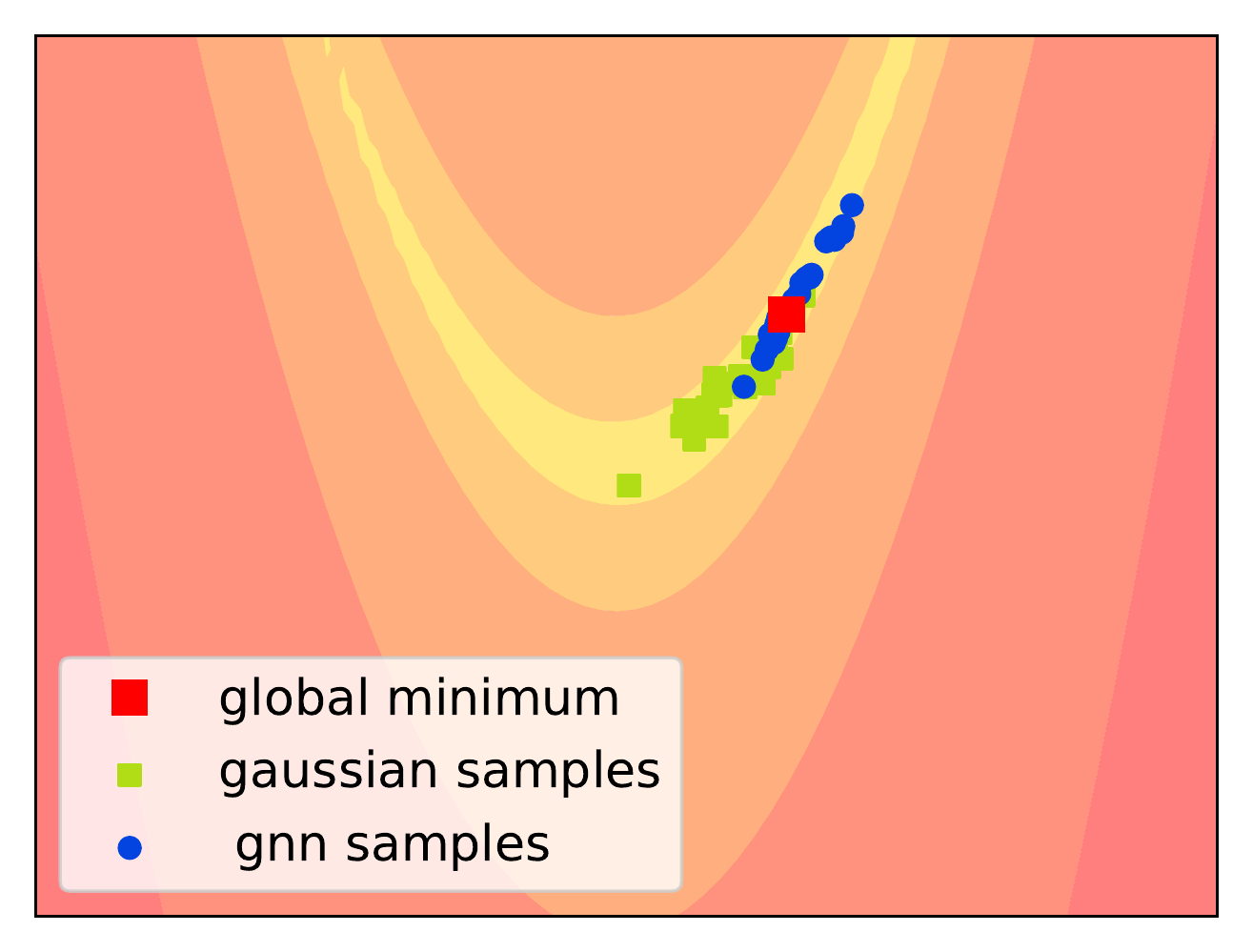}
    \caption{Iteration 15}
\end{subfigure}
 \caption{Evolution of distribution trained by GNN-xNES and xNES on the Rosenbrock function.}
 \label{fig::rosenbrock_gnn_samples}
\end{figure}

Figure \ref{fig::rosenbrock_gnn_samples} shows two iterations of the different algorithms on the Rosenbrock function, that typically highlights the Gaussian distribution lack of flexibility. On this function, GNN-xNES efficiently detects the curved valley and reaches the global minimum faster than xNES, while keeping a higher entropy. 

\begin{figure}
\begin{subfigure}{0.49\linewidth}
    \centering
    \includegraphics[width=\linewidth]{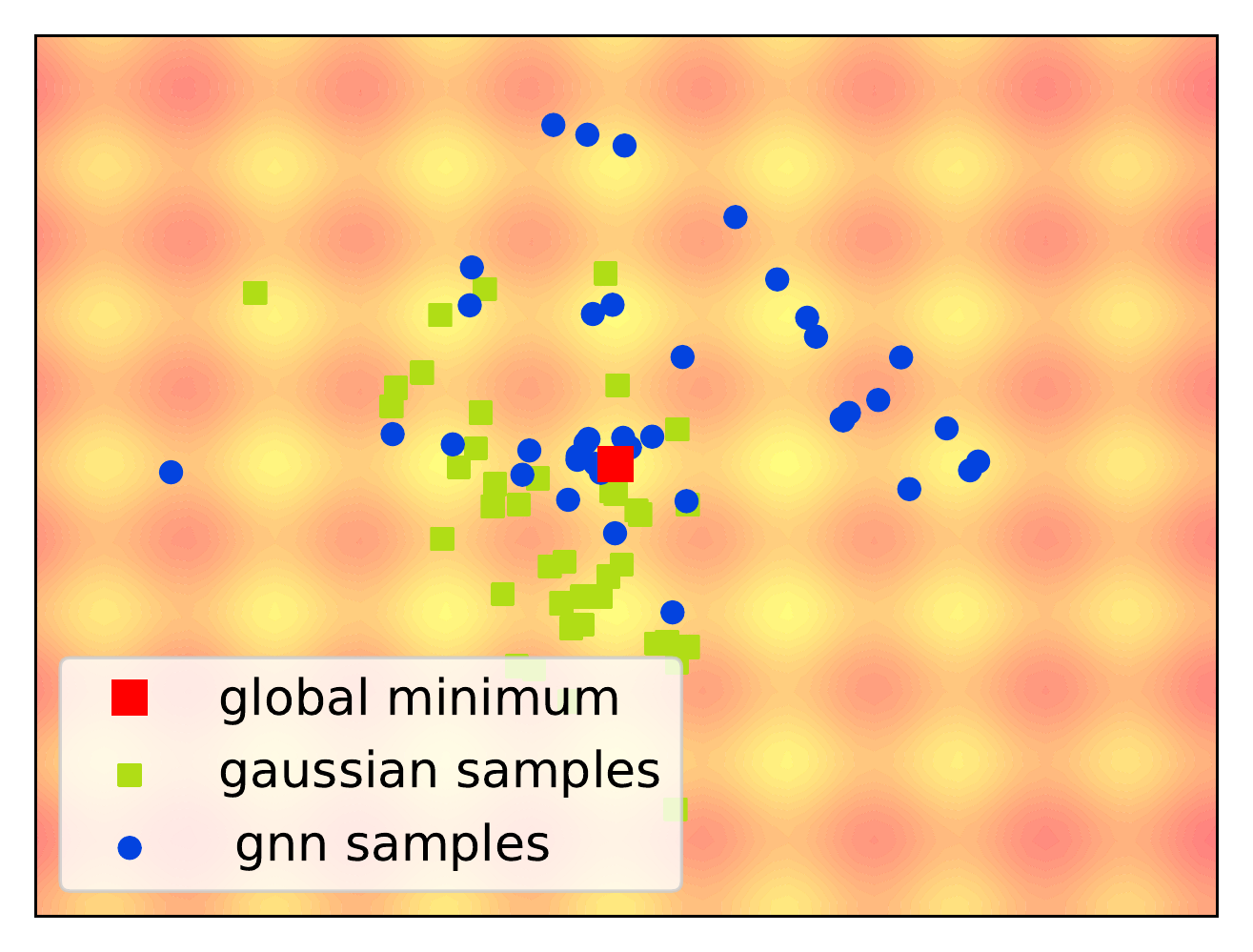}
    \caption{Iteration 10}
\end{subfigure}
\begin{subfigure}{0.49\linewidth}
    \centering
    \includegraphics[width=\linewidth]{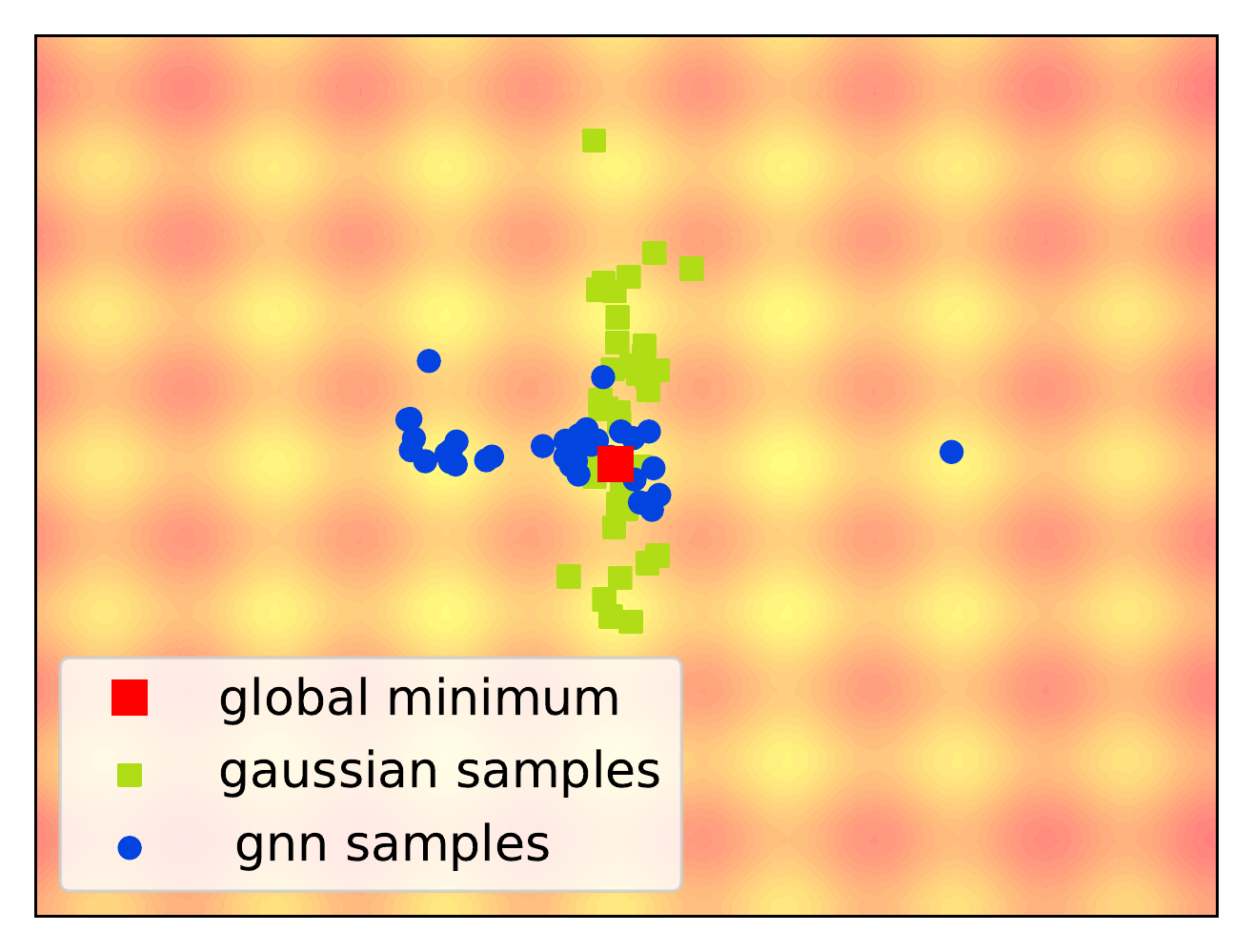}
    \caption{Iteration 20}
\end{subfigure}
 \caption{Evolution of distribution trained by GNN-xNES and xNES on the Rastrigin function.}
 \label{fig::rastrigin_gnn_samples}
\end{figure}

Figure \ref{fig::rastrigin_gnn_samples} provide a similar visualization on the Rastrigin function, a highly multimodal but symmetric objective. While both distributions discover the global minimum, GNN-xNES naturally creates a multi-model distribution, simultaneously exploring several local minima. 

\subsection{Synthetic objectives}

\paragraph{Experimental set-up}  We present experiments on both unimodal and multimodal optimization landscapes. We use official implementations\footnote{Implementation of xNES and CMA-ES were taken respectively from the PyBrain \cite{pybrain2010jmlr} library, and the PyCMA package available at \texttt{https://github.com/CMA-ES/pycma}.} of xNES and CMA-ES as baselines and as inner optimization methods of GNN-xNES and GNN-CMA-ES. Both xNES and CMA-ES are used with their default (adaptive) hyper-parameters.

For all synthetic objectives, the optimization landscape is randomly translated in the compact $[-2, 2]^d$ at the beginning of every ES trajectory to evaluate multiple configurations with different global minimum positions.
All algorithms are benchmarked on the same translated version of the objective, with similar initialization of their respective search distributions.

We build the NICE model with three coupling layers. Each coupling layer's non-linear mapping $t$ is built with a one hidden layer MLP, with 16 neurons and hyperbolic tangent activation. This architecture is kept constant in all our experiments. Other parameters, like the Kullback-Leibler radius $\varepsilon$ is also kept constant in all experiments, with value $\varepsilon=0.01$.

\paragraph{Unimodal landscapes}
We run the different algorithms on asymmetric landscapes, where we expect GNN search distributions to bring a significant improvement compared to the Gaussian - as discussed in \ref{subsec::limit}. These objectives functions are the Rosenbrock function (a curved valley with high conditioning) and the Bent Cigar function (a slightly better conditioned curved valley). Those two landscapes are often used in ES benchmarks \cite{hansen2010real} to evaluate a search distribution abilities to continuously adapt its search direction. 

We report performance measured in terms of the best current found value of the objective as a function of the number of function evaluations. Results are averaged over 10 random seeds. The population size is set to $N=10d$ for all algorithms.

\begin{figure}
    \begin{subfigure}{0.49\linewidth}
        \centering
        \includegraphics[width=\linewidth]{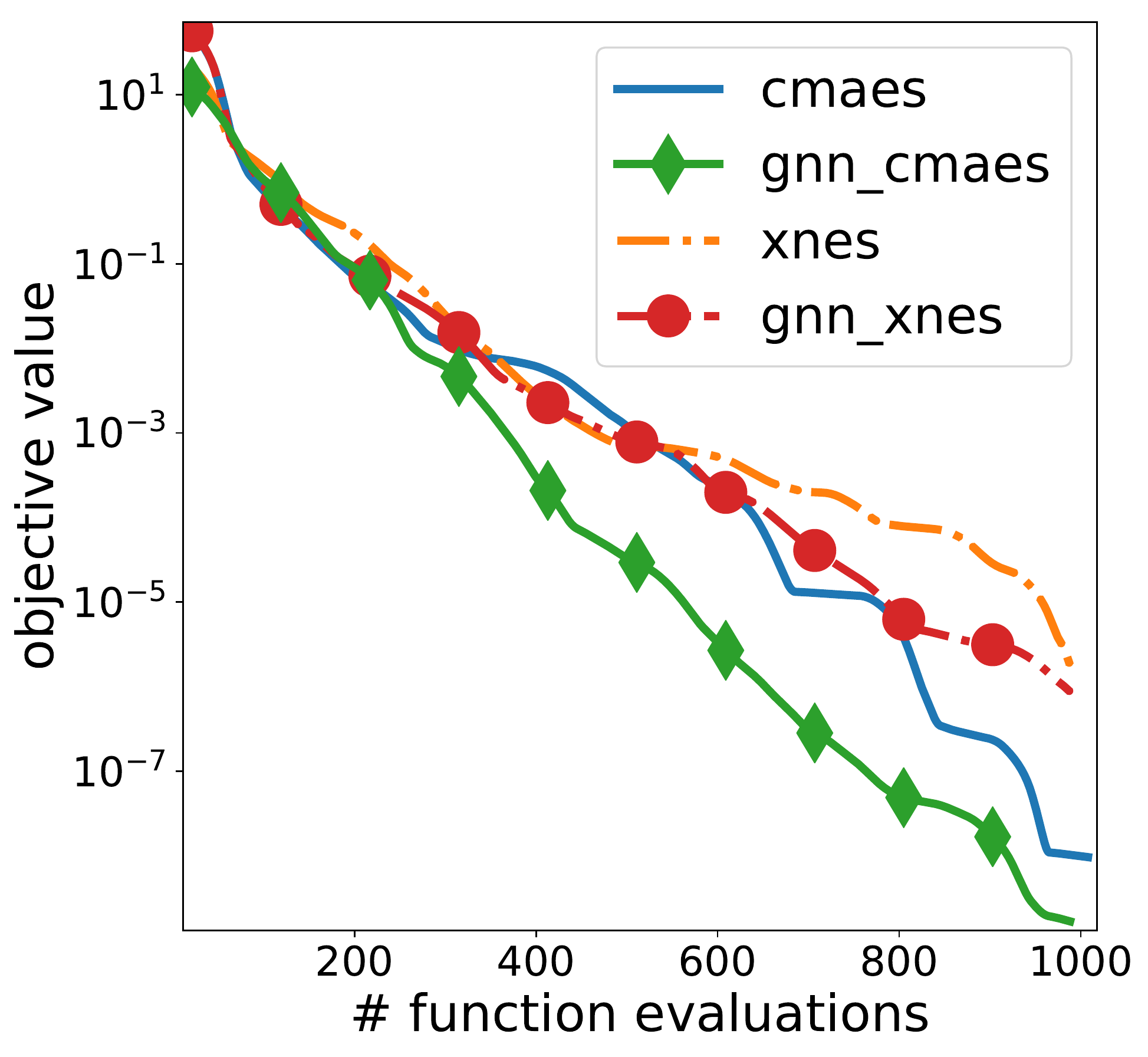}
        \caption{Bent Cigar, d=2}
        \label{fig:Bent_Cigar_d2}
    \end{subfigure}
    \begin{subfigure}{0.49\linewidth}
        \centering
        \includegraphics[width=\linewidth]{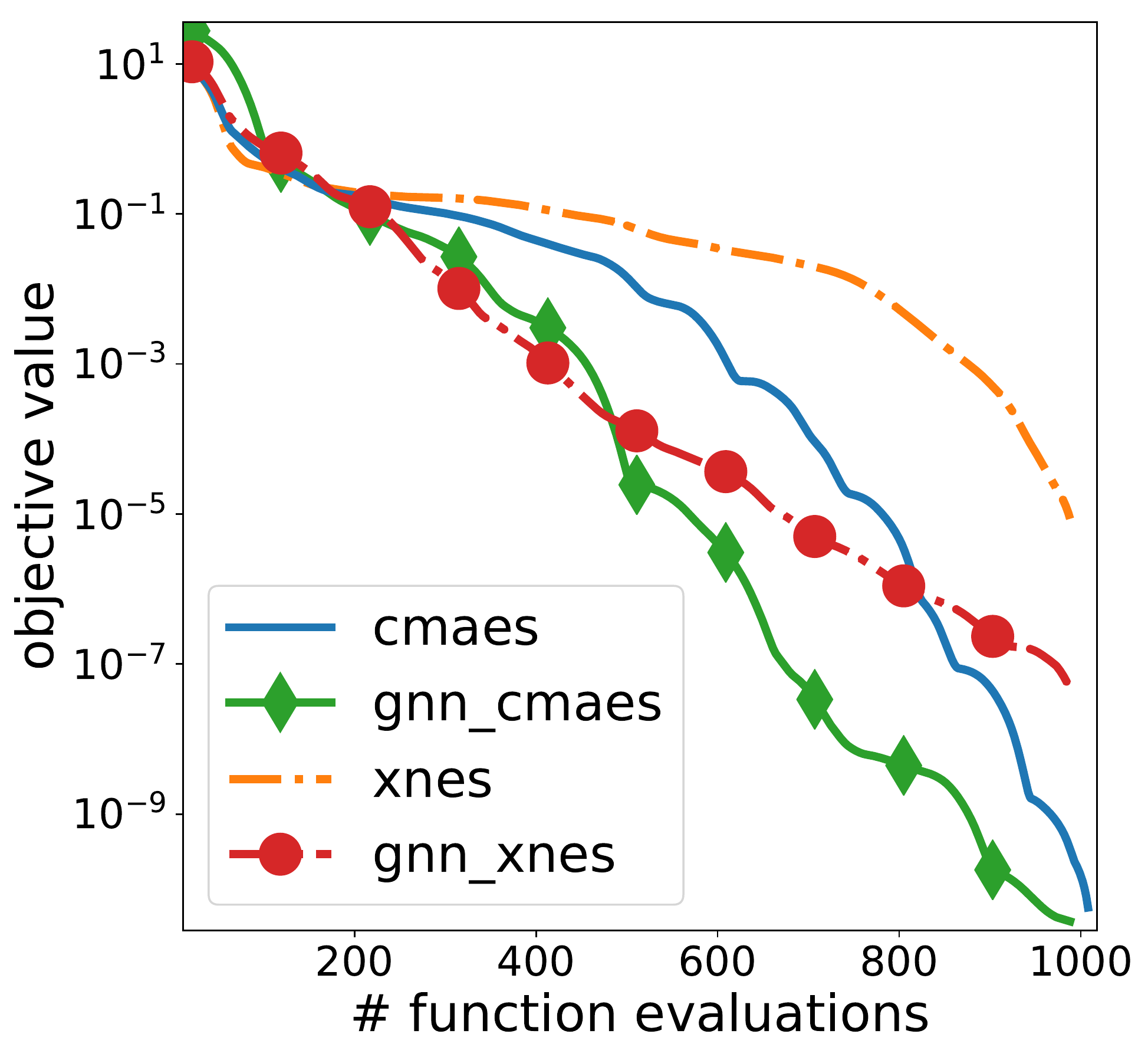}
        \caption{Rosenbrock, d=2}
        \label{fig:Rosenbrock_d2}
    \end{subfigure}\\
        \begin{subfigure}{0.49\linewidth}
        \centering
        \includegraphics[width=\linewidth]{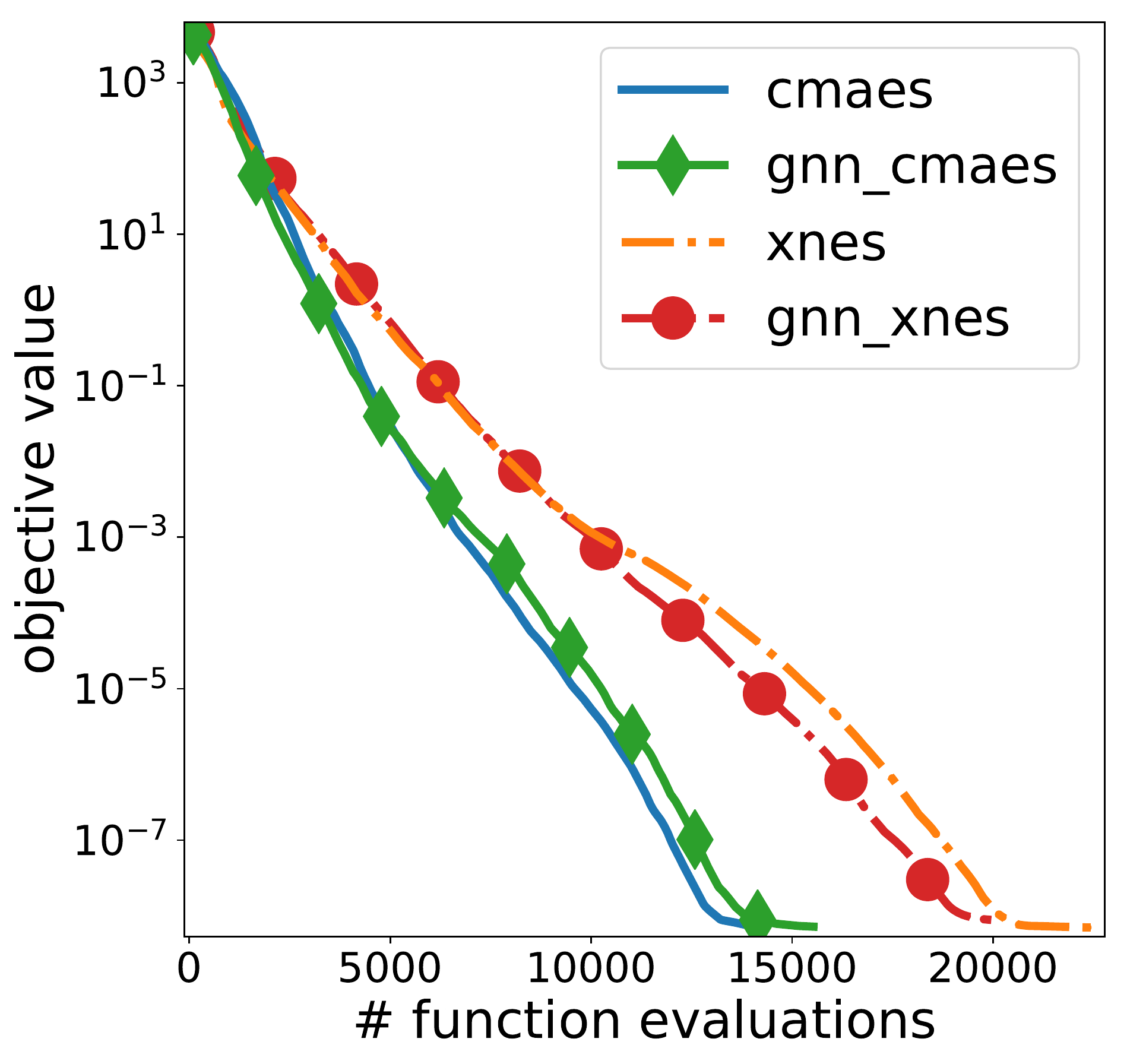}
        \caption{Bent Cigar, d=10}
        \label{fig:Bent_Cigar_d10}
    \end{subfigure}
    \begin{subfigure}{0.49\linewidth}
        \centering
        \includegraphics[width=\linewidth]{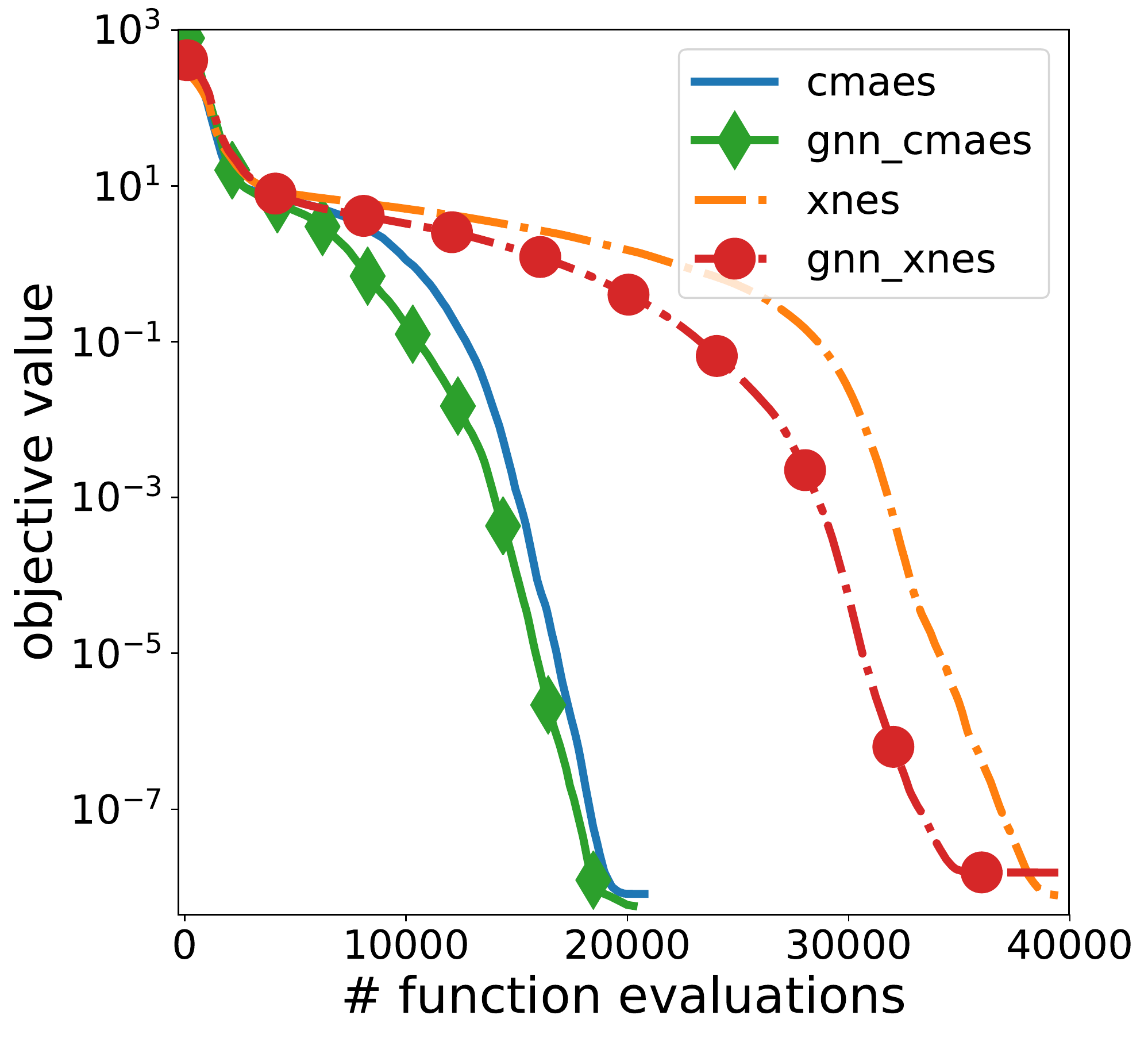}
        \caption{Rosenbrock, d=10}
        \label{fig:Rosenbrock_d10}
    \end{subfigure}
    \caption{Unimodal experiments}
    \label{fig::unimodal_experiments}
\end{figure}

Results for $d=2$ are presented in Figure \ref{fig:Bent_Cigar_d2} and \ref{fig:Rosenbrock_d2}. Systematically, the GNN models (GNN-CMA-ES and GNN-xNES) significantly outperform their respective baselines CMA-ES and xNES by achieving a faster convergence to the minimum. The mean improvement is quite significant, as at a given number of function evaluations the performance uplift can be of several orders of magnitude.

%The model GNN-CMAES reaches an improvement ratio up to $+10\%$  for \ref{fig:Bent_Cigar_d2} and \ref{fig:Rosenbrock_d2} around function evaluation $t \in (300,800)$. For xNES \ref{fig:Bent_Cigar_d2} the model achieve $+5\%$ around $800$  and  $+20\%$ for $t \in (600,800)$ for \ref{fig:Rosenbrock_d2}.

Results in higher dimensions ($d=10$) are presented in Figure \ref{fig:Bent_Cigar_d10} and \ref{fig:Rosenbrock_d10}. On the Bent Cigar function, we observe that GNN-CMA-ES does not improve over CMA-ES. We attribute this behavior to faster concentration of CMA-ES near the global optimum (the performances slightly diverge only close to the minimum).On the other hand, GNN-xNES strictly improved over xNES on this landscape. On the Rosenbrock function, both the GNN-based algorithm improve over their respective baselines. On this landscape, the speed-up is quite significant, as it reaches orders of magnitude at the beginning of the ES procedure. For instance, at $10^4$ function evaluations, GNN-CMA-ES discovers objective values approximately 10 time better than its counterpart CMA-ES.

\paragraph{Multimodal landscapes}
We report the performances of the different algorithms on multimodal objectives in Table \ref{tab:multimodal}. We measure performance as the value of the minimum (possibly local) of the objective function discovered by the ES algorithms at convergence, averaged over 20 random seeds. We consider four multimodal landscapes with different structures: the Rastrigin, the Styblinski, the Griewank and the Beale functions. Their expressions are reported in Appendix~\ref{sec::appendix}.

\begin{table}
    \centering
        \caption{Multimodal results}
        \vskip 0.1in
    \begin{tabular}{c|cccc|}
         \textbf{Objective} & \multicolumn{4}{|c|}{\textbf{Average Minimum Value}}  \\
         & \tiny{CMA-ES} & \tiny{GNN-CMA-ES} & \tiny{XNES} & \tiny{GNN-XNES} \\
         \footnotesize{Styblinski d=2} &  \footnotesize{11.3} &  \footnotesize{\textbf{5.65}} &  \footnotesize{9.89} &  \footnotesize{9.89}   \\
         \footnotesize{Styblinski d=4} &  \footnotesize{25.44} &  \footnotesize{\textbf{14.1}} &  \footnotesize{29.68} &  \footnotesize{18.37}   \\
         \footnotesize{Rastrigin d=2} &  \footnotesize{1.03} &  \footnotesize{1.14} &  \footnotesize{\textbf{0.29}} &  \footnotesize{0.49} \\
         \footnotesize{Rastrigin d=4} &  \footnotesize{1.60} &  \footnotesize{2.82} &  \footnotesize{\textbf{0.65}} &  \footnotesize{3.56} \\
         \footnotesize{Griewank d=2} &  \footnotesize{0.005} &  \footnotesize{\textbf{0.003}} &  \footnotesize{0.005} &  \footnotesize{0.025} \\
         \footnotesize{Griewank d=4} &  \footnotesize{0.002} &  \footnotesize{\textbf{0.001}} &  \footnotesize{0.003} &  \footnotesize{0.003} \\
          \footnotesize{Beale, d=2} &  \footnotesize{0.14} &  \footnotesize{\textbf{0.05}} &  \footnotesize{0.09} &  \footnotesize{0.09}   \\
          \footnotesize{Beale, d=4} &  \footnotesize{0.10} &  \footnotesize{\textbf{0.06}} &  \footnotesize{0.34} &  \footnotesize{0.09}   \\
    \end{tabular}
    \label{tab:multimodal}
\end{table}

GNN-CMA-ES constantly discovers better global minimum than other algorithms, except for one test function, the Rastrigin. This objective is highly multimodal and with poor global structure (i.e relatively \emph{flat}). Apart from leveraging its symmetry, the Gaussian search distribution has less capacity and is less likely to prematurely converge to a local minima in such a highly multimodal landscape - which explains its better results in this case. The benefits of using GNN are more highlighted in the other functions which have less local minima and better global structure, which allows the GNN search distribution to efficiently explore several of them at the same time and to discover more often the global minimum.

\subsection{Reinforcement Learning experiments}
ES algorithms have recently been used for direct policy search in Reinforcement Learning (RL) and shown to reach performances comparable with state-of-the-art MDP-based techniques \cite{liu2019trust, salimans2017evolution}. Direct Policy Search forgets the MDP structure of the RL environment and rather considers it as a \emph{black-box}. The search for the optimal policy is performed directly in parameter space to maximize the average reward per trajectory:
\begin{equation}
    f(x) = \lE_{\tau\sim p_x} \left[ \sum_{j\in\tau} r_j\right]
    \label{eq::reward_fun}
\end{equation}
where $p_x$ is the distribution of trajectories induced by the policy (the state-conditional distribution over actions) parametrized by $x$, and $r$ the rewards generated by the environment. 

The objective \eqref{eq::reward_fun} can readily be approximated from samples by simply rolling out $M$ trajectories, and optimized using ES. In our experiments, we set $M=10$ and optimize deterministic linear policies\footnote{We used the \texttt{rllab} library \cite{duan2016benchmarking} for experiments.}.

\begin{figure}
    \centering
     \begin{subfigure}{0.49\linewidth}
        \includegraphics[width=\linewidth]{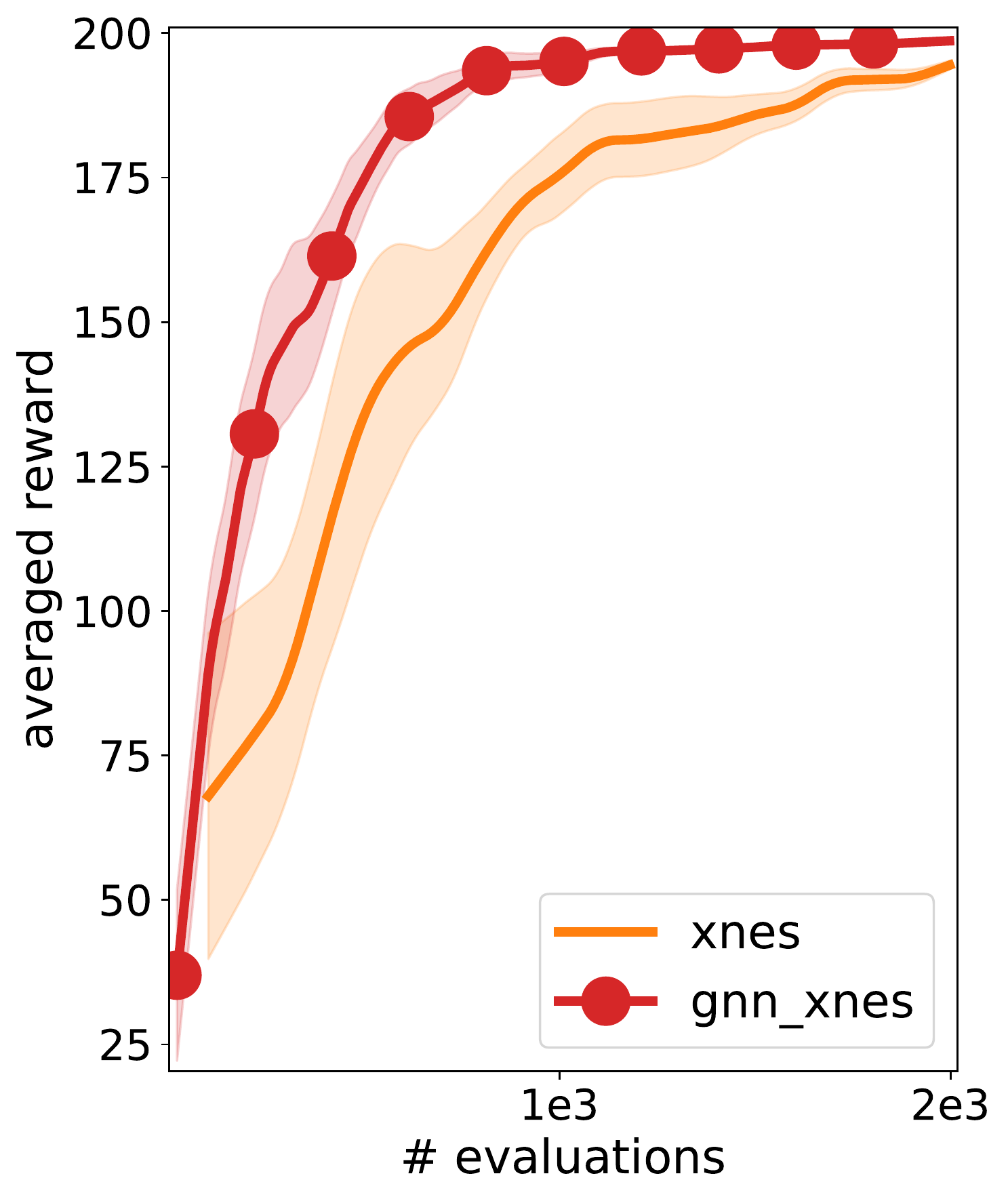}
        \caption{Swimmer-v1}
        \label{fig::rl_lunar}
    \end{subfigure}
     \begin{subfigure}{0.49\linewidth}
        \includegraphics[width=\linewidth]{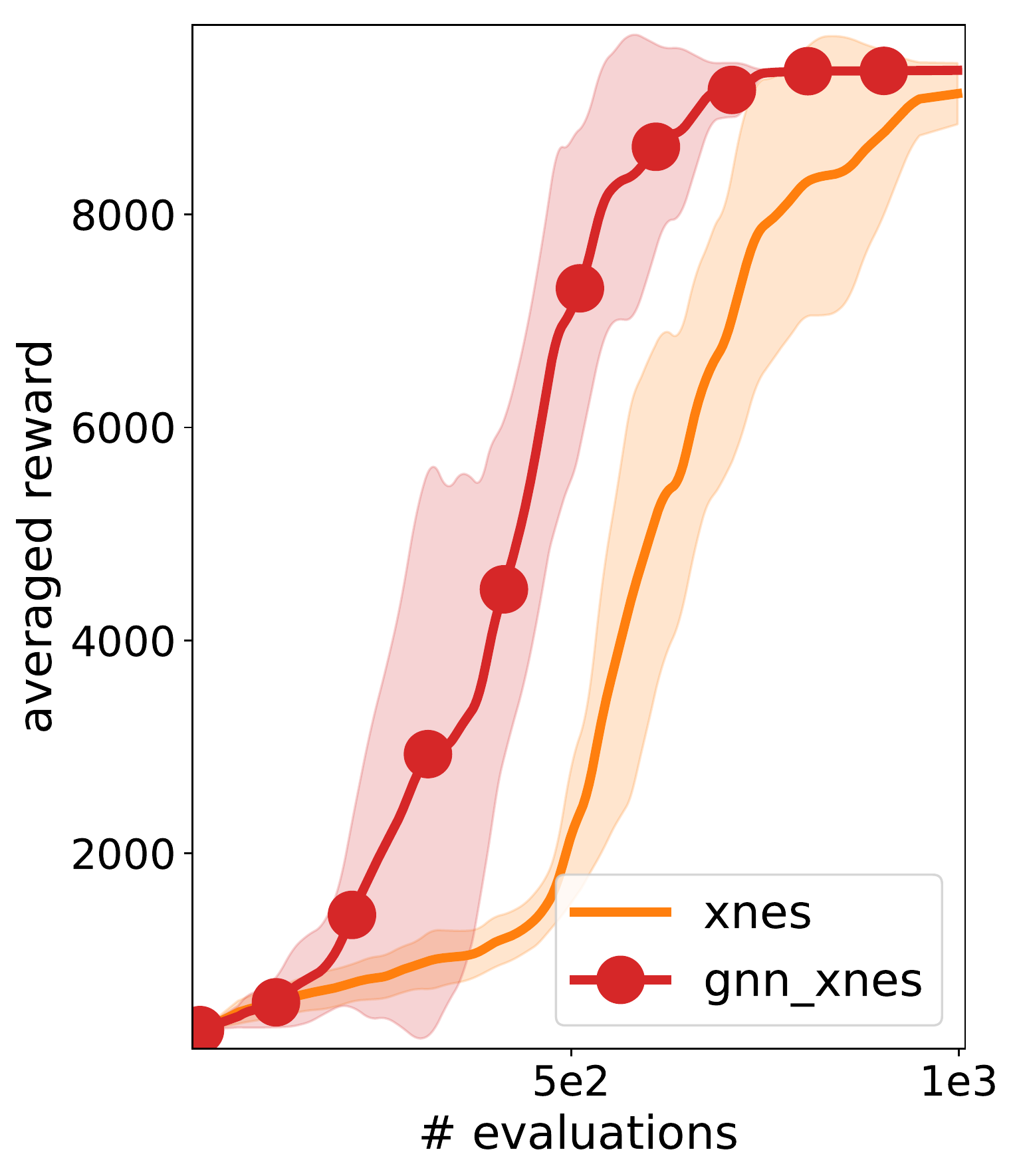}
        \caption{InvertedDoublePendulum}
        \label{fig::rl_swimmer}
    \end{subfigure}
    \caption{Direct Policy Search experiments}
\end{figure}

In Figures \ref{fig::rl_lunar} and \ref{fig::rl_swimmer} we report results of the GNN-xNES algorithm compared to xNES, when run on the Mujoco locomotion tasks Swimmer And InvertedDoublePendulum, both from the OpenAI Gym \cite{brockman2016openai}. Performance is measured  by the average reward per trajectory and per ES populations as a function of the number of evaluations of the objective $f$. Results are averaged over 5 random seeds (ruling the initialization of the environment and the initial distribution over the policy parameters $x$). In both environments, GNN-xNES discovers behaviors of high rewards faster than xNES. 

\section{Conclusion}
In this work, we motivate the use of GNNs for improving Evolutionary Strategies by pinpointing the limitations of classical search distributions, commonly used by standard ES algorithms. We propose a new algorithm that leverages the high flexibility of distributions generated by bijective GNNs with an ES objective. We highlight that this algorithm can be seen as a plug-in extension to existing ES algorithms, and therefore can virtually incorporate \emph{any} of them. Finally, we show its empirical advantages across a diversity of synthetic objective functions, as well as from objectives coming from Reinforcement Learning.

As possible extensions to our method, we could investigate \emph{noisy} zeroth-order oracles as well as first-order oracles, a problem already explored in \cite{grathwohl2017backpropagation, faury2018neural}.

\bibliography{bib}

\begin{thebibliography}{34}
\providecommand{\natexlab}[1]{#1}
\providecommand{\url}[1]{\texttt{#1}}
\expandafter\ifx\csname urlstyle\endcsname\relax
  \providecommand{\doi}[1]{doi: #1}\else
  \providecommand{\doi}{doi: \begingroup \urlstyle{rm}\Url}\fi

\bibitem[Agarwal et~al.(2017)Agarwal, Basu, Schnabel, and
  Joachims]{agarwal2017effective}
Agarwal, A., Basu, S., Schnabel, T., and Joachims, T.
\newblock Effective evaluation using logged bandit feedback from multiple
  loggers.
\newblock In \emph{Proceedings of the 23rd ACM SIGKDD International Conference
  on Knowledge Discovery and Data Mining}, pp.\  687--696. ACM, 2017.

\bibitem[Amari(1998)]{amari1998natural}
Amari, S.-I.
\newblock Natural gradient works efficiently in learning.
\newblock \emph{Neural {C}omputation}, 10\penalty0 (2):\penalty0 251--276,
  1998.

\bibitem[Brockman et~al.(2016)Brockman, Cheung, Pettersson, Schneider,
  Schulman, Tang, and Zaremba]{brockman2016openai}
Brockman, G., Cheung, V., Pettersson, L., Schneider, J., Schulman, J., Tang,
  J., and Zaremba, W.
\newblock Open{A}i {G}ym.
\newblock \emph{arXiv preprint arXiv:1606.01540}, 2016.

\bibitem[Dinh et~al.(2014)Dinh, Krueger, and Bengio]{dinh2014nice}
Dinh, L., Krueger, D., and Bengio, Y.
\newblock {{NICE}: {N}on-{L}inear {I}ndependent {C}omponents {E}stimation}.
\newblock \emph{arXiv preprint arXiv:1410.8516}, 2014.

\bibitem[Dinh et~al.(2016)Dinh, Sohl-Dickstein, and Bengio]{dinh2016density}
Dinh, L., Sohl-Dickstein, J., and Bengio, S.
\newblock {Density {E}stimation using {R}eal {NVP}}.
\newblock \emph{arXiv preprint arXiv:1605.08803}, 2016.

\bibitem[Duan et~al.(2016)Duan, Chen, Houthooft, Schulman, and
  Abbeel]{duan2016benchmarking}
Duan, Y., Chen, X., Houthooft, R., Schulman, J., and Abbeel, P.
\newblock {Benchmarking {D}eep {R}einforcement {L}earning for {C}ontinuous
  {C}ontrol}.
\newblock In \emph{International Conference on Machine Learning}, pp.\
  1329--1338, 2016.

\bibitem[Faury et~al.(2018)Faury, Vasile, Calauz{\`e}nes, and
  Fercoq]{faury2018neural}
Faury, L., Vasile, F., Calauz{\`e}nes, C., and Fercoq, O.
\newblock Neural generative models for global optimization with gradients.
\newblock \emph{arXiv preprint arXiv:1805.08594}, 2018.

\bibitem[Friedrichs \& Igel(2005)Friedrichs and
  Igel]{friedrichs2005evolutionary}
Friedrichs, F. and Igel, C.
\newblock Evolutionary tuning of multiple svm parameters.
\newblock \emph{Neurocomputing}, 64:\penalty0 107--117, 2005.

\bibitem[Goodfellow et~al.(2014)Goodfellow, Pouget-Abadie, Mirza, Xu,
  Warde-Farley, Ozair, Courville, and Bengio]{goodfellow2014generative}
Goodfellow, I., Pouget-Abadie, J., Mirza, M., Xu, B., Warde-Farley, D., Ozair,
  S., Courville, A., and Bengio, Y.
\newblock {Generative Adversarial Nets}.
\newblock In \emph{Advances in Neural Information Processing Systems}, pp.\
  2672--2680, 2014.

\bibitem[Grathwohl et~al.(2017)Grathwohl, Choi, Wu, Roeder, and
  Duvenaud]{grathwohl2017backpropagation}
Grathwohl, W., Choi, D., Wu, Y., Roeder, G., and Duvenaud, D.
\newblock Backpropagation through the void: Optimizing control variates for
  black-box gradient estimation.
\newblock \emph{arXiv preprint arXiv:1711.00123}, 2017.

\bibitem[Hansen(2016)]{hansen2016cma}
Hansen, N.
\newblock The {CMA} {E}volution {S}trategy: a tutorial.
\newblock \emph{arXiv preprint arXiv:1604.00772}, 2016.

\bibitem[Hansen \& Ostermeier(2001)Hansen and Ostermeier]{hansen2001completely}
Hansen, N. and Ostermeier, A.
\newblock Completely derandomized self-adaptation in {E}volution {S}trategies.
\newblock \emph{Evolutionary {C}omputation}, 9\penalty0 (2):\penalty0 159--195,
  2001.

\bibitem[Hansen et~al.(2010)Hansen, Auger, Finck, and Ros]{hansen2010real}
Hansen, N., Auger, A., Finck, S., and Ros, R.
\newblock \emph{Real-parameter black-box optimization benchmarking 2010:
  Experimental setup}.
\newblock PhD thesis, {INRIA}, 2010.

\bibitem[Jones et~al.(1998)Jones, Schonlau, and Welch]{jones1998efficient}
Jones, D.~R., Schonlau, M., and Welch, W.~J.
\newblock Efficient global optimization of expensive black-box functions.
\newblock \emph{Journal of Global optimization}, 13\penalty0 (4):\penalty0
  455--492, 1998.

\bibitem[Kingma \& Welling(2013)Kingma and Welling]{kingma2013auto}
Kingma, D.~P. and Welling, M.
\newblock Auto-encoding variational bayes.
\newblock \emph{arXiv preprint arXiv:1312.6114}, 2013.

\bibitem[Liu et~al.(2019)Liu, Zhao, Yang, Bian, Qin, Yu, and Liu]{liu2019trust}
Liu, G., Zhao, L., Yang, F., Bian, J., Qin, T., Yu, N., and Liu, T.-Y.
\newblock {T}rust {R}egion {E}volution {S}trategies.
\newblock 2019.

\bibitem[Loshchilov \& Hutter(2016)Loshchilov and Hutter]{loshchilov2016cma}
Loshchilov, I. and Hutter, F.
\newblock {CMA-ES} for hyperparameter optimization of deep neural networks.
\newblock \emph{arXiv preprint arXiv:1604.07269}, 2016.

\bibitem[MacKay(1995)]{mackay1995bayesian}
MacKay, D.~J.
\newblock Bayesian neural networks and density networks.
\newblock \emph{Nuclear Instruments and Methods in Physics Research Section A:
  Accelerators, Spectrometers, Detectors and Associated Equipment},
  354\penalty0 (1):\penalty0 73--80, 1995.

\bibitem[Martens(2010)]{martens2010deep}
Martens, J.
\newblock {Deep {L}earning via {H}essian-free {O}ptimization}.
\newblock In \emph{International {C}onference of {M}achine {L}earning},
  volume~27, pp.\  735--742, 2010.

\bibitem[Nedelec et~al.(2017)Nedelec, Roux, and
  Perchet]{nedelec2017comparative}
Nedelec, T., Roux, N.~L., and Perchet, V.
\newblock A comparative study of counterfactual estimators.
\newblock \emph{arXiv preprint arXiv:1704.00773}, 2017.

\bibitem[Nocedal \& Wright(2006)Nocedal and Wright]{NoceWrig06}
Nocedal, J. and Wright, S.~J.
\newblock \emph{Numerical Optimization}.
\newblock Springer, New York, NY, USA, second edition, 2006.

\bibitem[Rechenberg(1978)]{rechenberg1978evolutionsstrategien}
Rechenberg, I.
\newblock Evolutionsstrategien.
\newblock In \emph{Simulationsmethoden in der Medizin und Biologie}, pp.\
  83--114. Springer, 1978.

\bibitem[Rippel \& Adams(2013)Rippel and Adams]{rippel2013high}
Rippel, O. and Adams, R.~P.
\newblock High-dimensional {P}robability {E}stimation with {D}eep {D}ensity
  {M}odels.
\newblock \emph{arXiv preprint arXiv:1302.5125}, 2013.

\bibitem[Salimans et~al.(2017)Salimans, Ho, Chen, Sidor, and
  Sutskever]{salimans2017evolution}
Salimans, T., Ho, J., Chen, X., Sidor, S., and Sutskever, I.
\newblock Evolution {S}trategies as a scalable alternative to {R}einforcement
  {L}earning.
\newblock \emph{arXiv preprint arXiv:1703.03864}, 2017.

\bibitem[Schaul et~al.(2010)Schaul, Bayer, Wierstra, Sun, Felder, Sehnke,
  R{\"u}ckstie{\ss}, and Schmidhuber]{pybrain2010jmlr}
Schaul, T., Bayer, J., Wierstra, D., Sun, Y., Felder, M., Sehnke, F.,
  R{\"u}ckstie{\ss}, T., and Schmidhuber, J.
\newblock {PyBrain}.
\newblock \emph{Journal of Machine Learning Research}, 11:\penalty0 743--746,
  2010.

\bibitem[Schaul et~al.(2011)Schaul, Glasmachers, and
  Schmidhuber]{schaul2011high}
Schaul, T., Glasmachers, T., and Schmidhuber, J.
\newblock {High dimensions and heavy tails for {N}atural {E}volution
  {S}trategies}.
\newblock In \emph{Proceedings of the 13th annual conference on {G}enetic and
  {E}volutionary {C}omputation}, pp.\  845--852. ACM, 2011.

\bibitem[Schulman et~al.(2017)Schulman, Wolski, Dhariwal, Radford, and
  Klimov]{schulman2017proximal}
Schulman, J., Wolski, F., Dhariwal, P., Radford, A., and Klimov, O.
\newblock Proximal policy optimization algorithms.
\newblock \emph{arXiv preprint arXiv:1707.06347}, 2017.

\bibitem[Schwefel(1977)]{schwefel1977numerische}
Schwefel, H.-P.
\newblock \emph{Numerische Optimierung von Computer-Modellen mittels der
  Evolutionsstrategie: mit einer vergleichenden Einf{\"u}hrung in die
  Hill-Climbing-und Zufallsstrategie}.
\newblock Birkh{\"a}user, 1977.

\bibitem[Shahriari et~al.(2016)Shahriari, Swersky, Wang, Adams, and
  De~Freitas]{shahriari2016taking}
Shahriari, B., Swersky, K., Wang, Z., Adams, R.~P., and De~Freitas, N.
\newblock Taking the human out of the loop: A review of {B}ayesian
  {O}ptimization.
\newblock \emph{Proceedings of the IEEE}, 104\penalty0 (1):\penalty0 148--175,
  2016.

\bibitem[Srivastava et~al.(2017)Srivastava, Valkoz, Russell, Gutmann, and
  Sutton]{srivastava2017veegan}
Srivastava, A., Valkoz, L., Russell, C., Gutmann, M.~U., and Sutton, C.
\newblock Veegan: Reducing mode collapse in {GAN}s using {I}mplicit
  {V}ariational {L}earning.
\newblock In \emph{Advances in Neural Information Processing Systems}, pp.\
  3308--3318, 2017.

\bibitem[Sun et~al.(2009)Sun, Wierstra, Schaul, and
  Schmidhuber]{sun2009efficient}
Sun, Y., Wierstra, D., Schaul, T., and Schmidhuber, J.
\newblock Efficient {N}atural {E}volution {S}trategies.
\newblock In \emph{Proceedings of the 11th {A}nnual {C}onference on {G}enetic
  and {E}volutionary {C}omputation}, pp.\  539--546. ACM, 2009.

\bibitem[Swaminathan \& Joachims(2015)Swaminathan and
  Joachims]{swaminathan2015counterfactual}
Swaminathan, A. and Joachims, T.
\newblock Counterfactual {R}isk {m}inimization: Learning from {L}ogged {B}andit
  {F}eedback.
\newblock In \emph{International Conference on Machine Learning}, pp.\
  814--823, 2015.

\bibitem[Wierstra et~al.(2008)Wierstra, Schaul, Peters, and
  Schmidhuber]{wierstra2008natural}
Wierstra, D., Schaul, T., Peters, J., and Schmidhuber, J.
\newblock {Natural {E}volution {S}trategies}.
\newblock In \emph{Evolutionary {o}mputation, 2008. {CEC} 2008.({IEEE} {W}orld
  {C}ongress on {C}omputational {I}ntelligence)}, pp.\  3381--3387. IEEE, 2008.

\bibitem[Williams(1992)]{williams1992simple}
Williams, R.~J.
\newblock Simple statistical gradient-following algorithms for connectionist
  {R}einforcement {L}earning.
\newblock \emph{Machine Learning}, 8\penalty0 (3-4):\penalty0 229--256, 1992.

\end{thebibliography}
\bibliographystyle{icml2019}

\clearpage

\appendix
\section{Experimental details}
\label{sec::appendix}

\subsection{Synthetic objectives}
We provide in Table \ref{tab::expressions} the expressions of the synthetic objective functions used in this paper. We use the asymmetric operator $T_{\text{asy}}^\beta$ for the Bent Cigar function introduced in \cite{hansen2010real}. $R$ denotes a (random) rotation matrix in $\mathbb{R}^d$. We use $\beta=0.5$ for $d=2$ and $\beta=2$ for $d=10$.

\begin{table}[h]
    \centering
    \caption{Synthetic objectives}
    \begin{tabular}{l|l}
        \textbf{Objective function} & Expression \\ \hline
         Rosenbrock & $f_1(x) = \sum_i \left(1-x_i \right)^2 + 100 (x_{i+1}- x_i)$ \\ 
         Cigar & $f_2(x) = x_1^2 + 10^4 \sum_{i=2}^d x_i^2 $ \\
         Bent Cigar & $ f_3(x) =  f_2(RT_{\text{asy}}^\beta(x)R)$ \\
         Rastrigin & $f_4(x) = 10d + \sum_{i=1}^{d} \left(x_i^2 - A \cos \left( 2\pi x_i \right) \right)$ \\
         Griewank & $f_5(x) = \sum_{i=1}^d \frac{x_i^2}{4000}- \prod_{i=1}^{2} \cos \left(\frac{\sqrt{x_i}}{\sqrt{i}} \right) + 1 $ \\
         Beale &  $\begin{aligned}f_6(x) &= \left(1.5 - x_1 + x_1x_2\right)^2\\
                        &+\left(2.25 - x_1+x_1x_2^2\right)^2\\
                        &+ \left(2.625 - x_1 + x_1x_2^3\right)^2 + \sum_{i=3}^d x_i^2\end{aligned}$ \\
         Styblinski & $f_7(x) = \frac{1}{2} \sum_{i=1}^d \left(x_i^4 -16 x_i^2 + 5 x_i\right)$
    \end{tabular}
    \label{tab::expressions}
\end{table}

\subsection{RL environments}

Table \ref{tab:rl} provides details on the RL environment used to compare GNN-xNES and xNES, like the dimensions of the state space $\mathcal{S}$ and action space $\mathcal{A}$, and the maximum number of steps per trajectory. 

\begin{table}[h!]
    \caption{RL environment details}
    \centering
    \begin{tabular}{c|c|c|c}
          Name & $\vert \mathcal{S}\vert $ & $\vert\mathcal{A\vert}$ & Steps\\\hline
          Swimmer-v1  & 13 & 2 & 1000\\
          InvertedDoublePendulum-v1 & 11 & 1 & 1000
    \end{tabular}
    \label{tab:rl}
\end{table}

\end{document}